\documentclass[sigconf]{acmart}

\usepackage{amsmath,amsfonts} 
\usepackage{epsfig}
\usepackage{subfigure}
\usepackage{booktabs}
\usepackage{multicol}
\usepackage{url}
\usepackage{graphicx}

\begin{document}

\title{Cross-Platform Emoji Interpretation:\\Analysis, a Solution, and Applications}

\author{Fred Morstatter}
\affiliation{%
  \institution{University of Southern California}
  \streetaddress{4676 Admiralty Way Ste. 1001}
  \city{Marina Del Rey} 
  \state{California} 
  \postcode{90292}
}
\email{morstatt@usc.edu}

\author{Kai Shu}
\affiliation{%
  \institution{Arizona State University}
  \streetaddress{699 S. Mill Ave}
  \city{Tempe} 
  \state{Arizona} 
  \postcode{85283}
}
\email{kai.shu@asu.edu}

\author{Suhang Wang}
\affiliation{%
  \institution{Arizona State University}
  \streetaddress{699 S. Mill Ave}
  \city{Tempe} 
  \state{Arizona} 
  \postcode{85283}
}
\email{swang187@asu.edu}

\author{Huan Liu}
\affiliation{%
  \institution{Arizona State University}
  \streetaddress{699 S. Mill Ave}
  \city{Tempe} 
  \state{Arizona} 
  \postcode{85283}
}
\email{huan.liu@asu.edu}

\begin{abstract}
Most social media platforms are largely based on text, and users often write posts to describe where they are, what they are seeing, and how they are feeling. Because written text lacks the emotional cues of spoken and face-to-face dialogue, ambiguities are common in written language. This problem is exacerbated in the short, informal nature of many social media posts. To bypass this issue, a suite of special characters called ``emojis,'' which are small pictograms, are embedded within the text. Many emojis are small depictions of facial expressions designed to help disambiguate the emotional meaning of the text. However, a new ambiguity arises in the way that emojis are rendered. Every platform (Windows, Mac, and Android, to name a few) renders emojis according to their own style. In fact, it has been shown that some emojis can be rendered so differently that they look ``happy'' on some platforms, and ``sad'' on others. In this work, we use real-world data to verify the existence of this problem. We verify that the usage of the same emoji can be significantly different across platforms, with some emojis exhibiting different sentiment polarities on different platforms.  We propose a solution to identify the intended emoji based on the platform-specific nature of the emoji used by the author of a social media post. We apply our solution to sentiment analysis, a task that can benefit from the emoji calibration technique we use in this work. We conduct experiments to evaluate the effectiveness of the mapping in this task.
\end{abstract}

\maketitle

\section{Introduction}
Social media and web communication are a major part of every day life for most people. Sites like Facebook, Twitter, and WhatsApp all have hundreds of millions to billions of users who communicate on these platforms each and every day. While images and videos have become commonplace on these sites, text is still the predominant method of communication. This happens on our smartphones, tablets, and computers billions of times every day.

Communication through text has many key issues that keep it from having the depth of face-to-face conversation. One of these issues is the lack of emotional cues~\cite{daft1986organizational}. When conversation is carried out through text, the lack of non-verbal cues removes key emotional elements from the conversation~\cite{byron2005toward}. One solution to this problem is \emph{emoticons}, which are combinations of standard keyboard characters to create facial representations of human emotion, e.g., :), :(, \textasciicircum\_\textasciicircum, and :D. While widely used, there are only a limited number of character combinations that make a cogent representation of a human emotion, and the exact meaning of many emoticons can be ambiguous~\cite{walther2001impacts}. To provide for richer expression, ``emojis'' offer a richer set of non-verbal cues.

\emph{Emojis} are a set of reserved characters that, when rendered, are small pictograms that depict a facial expression, or other object. Unlike emoticons, these are not combinations of characters devised by the users, but instead single characters that are rendered as small pictures on the screen. There are currently over 1,800 different emojis defined by the Unicode specification, a number that grows with each iteration of the specification.\footnote{\url{http://www.unicode.org/Public/emoji/3.0/emoji-data.txt}} These emojis are either facial expressions (e.g., ``grinning face,'' \includegraphics[height=1em]{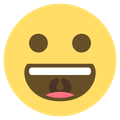}, character code U+1F600),\footnote{Unless otherwise noted, when an emoji appears inline, the depiction comes from Emoji One (\url{http://emojione.com/}), which is considered the standard.} or ideograms (e.g., ``birthday cake,'' \includegraphics[height=1em]{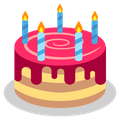}, character code U+1F382).

While emojis have allowed for increased expression of emotion through text, they have an inconsistency. That is, emojis are rendered differently on each platform. Just as different fonts display the same character according to a different style; similarly, each major platform has its own font to display these characters. With regular characters this is not crucial as each character has a predefined meaning. Emojis do not enjoy this predefined definition, and these changes in rendering can have an impact on the way that the emoji is interpreted. Emojis are often small depictions of human faces, so slight variations can make the face look entirely different. This can cause a different interpretation of the text than was initially intended by the author for emotional interpretation to the text. 
This issue was raised in Miller \emph{et al.}~\cite{miller2016blissfully}, where human workers on a crowdsourcing platform rated the sentiment of emojis. The results of these ratings indicate that the same emoji can be perceived as positive on some platforms, while it can be perceived as negative on others. Miller's finding is important, with repercussions for the 2 billion people who use a smartphone.\footnote{https://www.statista.com/statistics/330695/number-of-smartphone-users-worldwide/} Furthermore, recent research suggests that emojis are replacing emoticons on social media sites such as Twitter~\cite{pavalanathan2015emoticons}. With so many people affected by this possibility for miscommunication, it is important that we study the implications and possible solutions to this problem. 

In this paper, we answer the following research questions:
\begin{enumerate}
	\item[\textbf{RQ1}] \emph{Does misinterpretation based upon emoji rendering occur in real world data?} Miller \textit{et al.}~\cite{miller2016blissfully} discovered this possibility for misinterpretation using surveys. We assess if these phenomena appear in real world social media datasets. 
	\item[\textbf{RQ2}] \emph{What is the scale of this misinterpretation?} If this misinterpretation manifests, that does not necessarily mean that it affects a vast array of communication. We measure the extent to which communication on one social network, Twitter, is affected by misinterpretable emojis.
	\item[\textbf{RQ3}] \emph{How can the problem of cross-platform emoji interpretation be addressed?} Using our insights from the first two analytical portions, we construct a solution that produces a mapping of emojis from one platform to those on another. 
	\item[\textbf{RQ4}] \emph{Does correcting for emoji misinterpretation have a meaningful effect on analysis?} We measure the usefulness of our mapping by applying it to a common text analysis task: sentiment analysis. We show that the performance is increased by mapping all tweets to a common emoji language.
\end{enumerate}

\section{Related Work}
In this section we discuss the related work from three different perspectives.
First, since our solution heavily relies on word embeddings to create the emoji mapping, we enumerate some recent work on word embeddings.
Next, we discuss other resources for informal text and continue to discuss other work that has been done on emoji analysis.
Finally, we discuss other work that has been done in the context of platform-specific emoji rendering.

\subsection{Word Embeddings}
One of the first word embedding algorithms was the Neural Network Language Model~\cite{bengio2006neural}. Currently, one of the most famous word embedding algorithms, Word2Vec~\cite{mikolov2013distributed}, has risen to prominence. This algorithm works embedding words that appear next to each other in the text next to each other in the embedding. Word2Vec provides two approaches to solving this problem: the ``continuous bag of words'' (CBOW) and the ``skip gram'' (SG) architecture. While skip gram has been shown to better solve analogies~\cite{mikolov2013linguistic}, we compare with both approaches in this work.

The simple rules combined with other constraints has been shown to create powerful embeddings that work in many different settings. For example, they have been used to produce state-of-the-art performance in the areas of syntactic parsing~\cite{socher2013parsing}, named entity recognition~\cite{dhillon2011multi}, antonym detection~\cite{ono2015word}, sentiment analysis~\cite{maas2011learning,tang2014coooolll,tang2014learning}, and machine translation~\cite{zou2013bilingual}. 

\subsection{Linguistic Resources for Informal Text}
Before emojis were commonplace, there was a long history of trying to better represent and understand text, both formal and informal. One of the most influential resources is WordNet~\cite{miller1995wordnet}, which represents words not only by their definitions, but also provides a graph of the relationship between the words. Extensions to this approach abound, but perhaps the most relevant one to our work is SentiWordNet~\cite{esuli2006sentiwordnet}, which considers the sentiment of the words when building the resource. In the context of informal text, SlangSD~\cite{wu2016slangsd} provides a sentiment resource for mapping slang words to sentiment scores by leveraging Urban Dictionary's data. Crowdsourcing has also been used to extract the emotional meanings for words~\cite{mohammad2013crowdsourcing}.

\subsection{Emoji Analysis}
As emojis have become an important tool that help people communicate and express their emotions, the study of emojis as they pertain to sentiment classification and text understanding is attracting attention~\cite{barbieri2016cosmopolitan,barbieri2016does,eisner2016emoji2vec,hallsmarmulti,hu2013unsupervised,kelly2015characterising,novak2015sentiment}. Hu \emph{et al.}~\cite{hu2013unsupervised} proposes an unsupervised framework for sentiment classification by incorporating emoticon signals. Hallsmar \emph{et al.}~\cite{hallsmarmulti} investigates the feasibility of an emoji training heuristic for multi-class sentiment analysis on Twitter with a Multinomial Naive Bayes Classifier. Eisner \textit{et al.}~\cite{eisner2016emoji2vec} learn emoji representation by running skip gram on descriptions of emojis provided in the Unicode standard. Instead of using Unicode description, Barbieri \emph{et al.}~\cite{barbieri2016does} learns the vector skip gram model for twitter emojis using tweets, which also demonstrate the ability of Emojis in improving sentiment analysis. Others analyze the sentiment of emojis with respect to tweet corpus of different languages, position of emojis in text, etc (~\cite{barbieri2016cosmopolitan,novak2015sentiment}). We build on this work by studying the difference for emoji usage in different \emph{platforms}. 

\subsection{Platform Specific Emoji Rendering}
The aforementioned studies on emoji analysis ignore the fact that the same emoji unicode has different emoji images on different platform. Thus, the sentiment or semantic meanings of the same emoji may be perceived differently for people using different platforms and thus cause misunderstanding. Therefore, recently, there are researchers paying attention to this issue~\cite{miller2016blissfully,tigwell2016oh}.  Miller \textit{et al.}~\cite{miller2016blissfully} show that emoji misinterpretation exists within and across platforms, from both semantic and sentiment perspectives. The analysis is based on a survey to collect people's feedback of sentiment scores and semantic meaning on different rendering of emojis, which does not consider the context of emojis. Similarly, Tigwell \textit{et al.} also explore platform-dependent emoji misinterpretation problem in~\cite{tigwell2016oh}. They design a questionnaire to collect user's sentiment feedback on 16 emojis from Android and iOS platform, and compute a valence-arousal space to guide sentiment analysis. Different from existing approaches exploring platform specific emoji rendering problems, we use real world data to verify the existence of emoji ambiguity and provides a mapping-based solution to identify the intended emoji from original posts.

\section{Platform-Specific Emoji Usage}
Previous work by Miller \emph{et al.}~\cite{miller2016blissfully,miller2017understanding} identified platform-specific emoji meaning by carrying out surveys with human participants on Amazon's Mechanical Turk. While these insights are extremely useful, we must verify that these patterns truly occur in real-world data. In this section we outline our process for collecting an emoji dataset and measure the effect to which platform plays a role in the use of emojis.

\begin{table}
	\centering
	\caption{Amount of data collected by platform. The ``Source(s)'' column indicates the applications we chose to represent each platform.}
	\begin{tabular}{l l r}
		\toprule
		\textbf{Platform} & \textbf{Source(s)} & \textbf{Tweets} \\
		\midrule
		Android & Twitter for Android & 5,839,392\\
		& & \\
		iOS & Twitter for iPad & 12,850,344 \\
		& Twitter for iPhone & \\
		& iOS & \\
		& & \\
		Twitter & Twitter Web Client & 1,562,655 \\
		& & \\
		Windows &  Twitter for Windows Phone & 114,175 \\
		& Twitter for Windows & \\
		\midrule
		\emph{Total} & & 20,366,566 \\
		\bottomrule
	\end{tabular}
	\label{tab:data}
\end{table}

\subsection{A Platform-Specific Emoji Dataset}
The dataset used in this work consists of social media posts collected from Twitter. Twitter is an attractive option for our analysis for several reasons. First, it is large. With approximately 500 million tweets each day, it is one of the largest social media sites. Also, because of its 140-character limit, users may be prone to use emojis because they can help the user to be more expressive within the restrictive character limit. Furthermore, Twitter, like many other social networking sites, is a place where people post using many different platforms. Additionally, the site makes the source of the post available as part of its metadata.

To collect the dataset used in this work, we manually identify a subset of emojis that have human faces or other emotional signals.  The full list of codes used in the data collection will be released upon request. Using this list of emojis we query Twitter's Filter API,\footnote{https://dev.twitter.com/streaming/reference/post/statuses/filter} which takes as input a list of keywords to track, using our list of emojis. We tracked this data for 28 days, collecting a total of 20 million tweets.

Because the nature of this work is focused on the platform-specific nature of the emojis, we separate the tweets based on the platform from which they were posted. Twitter is an open platform, meaning that it has a fully documented and available API which any third party can use to make software to post to Twitter. While Twitter does not explicitly mention which platform the user used to write the tweet, they do provide details about the software used to author the tweet in their ``source'' field. This source is made available in the data that comes from their APIs. In some cases, the ``source'' is a clear indication of the platform because some software is only available on one platform. We use these when determining the platform from which the tweet was posted. We identify four platforms with distinct emoji sets according to Emojipedia:\footnote{\url{http://emojipedia.org/}} iOS, Android, Windows, and Twitter. We select these because they are major platforms. There are two reasons behind this. First, the results obtained from these platforms will apply to more users on the social media site. Second, by choosing large platforms we can accrue a more sizable dataset, which will yield a more stable mapping. Statistics of the dataset, as well as the applications we selected to represent each platform, are shown in Table~\ref{tab:data}.

\label{sec:dataanalysis}
Now that we have collected an emoji dataset, we will continue to investigate the differences between the usage of emojis on different platforms. Towards answering \textbf{RQ1}, this analysis is performed from two perspectives: 1) the positioning of the emojis within a word embedding, and 2) the sentiment of the posts in which the emojis appear.

\begin{table}
	\caption{Average Jaccard coefficient of emojis across platforms. Random indicates a random sample across all tweets of the size of the iOS corpus.}
	\resizebox{\columnwidth}{!}{
	\begin{tabular}{c c c c c c}
	\toprule
	& Android & iOS & Twitter & Windows & Random \\
	\midrule
	Android & 1.000 & 0.153 & 0.111 & 0.062 & 0.010 \\
	iOS & 0.153 & 1.000 & 0.086 & 0.052 & 0.018 \\
	Twitter & 0.111 & 0.086 & 1.000 & 0.047 & 0.005 \\
	Windows & 0.062 & 0.052 & 0.047 & 1.000 & 0.002 \\
	Random & 0.010 & 0.018 & 0.005 & 0.002 & 1.000 \\
	\bottomrule
	\end{tabular}}
	\label{tab:jaccard}
\end{table}
\subsection{Measuring Emoji Embedding Agreement}
First, we investigate how consistent the embeddings of the emojis are across platforms. Word2Vec~\cite{mikolov2013distributed}, a word embedding algorithm, learns a vector for each word in the vocabulary. For a given word, the vector is constructed based upon the neighboring words each time the given word is used. Thus, we employ this technique to measure how consistent the usage of emojis is across platforms.

We build a base word embedding by training a skip gram Word2Vec model using the entire dataset with emojis removed.\footnote{This process is explained in greater detail in Section~\ref{sec:mappingsolution}.} Then we use this base embedding to train a platform-specific embedding by adding the emojis back in and updating the model with the new data. It is important to note that we do not update non-emoji words, we only update the vectors of the newly-added emojis. After following this process, we have 4 platform-specific emoji embeddings. To measure how these deviate from general Twitter conversation, we also create a platform-agnostic word embedding by training a random sample of tweets of the size of the iOS platform.

\begin{figure*}
	\centering
	\hspace*{-.5cm}\includegraphics[width=.9\textwidth]{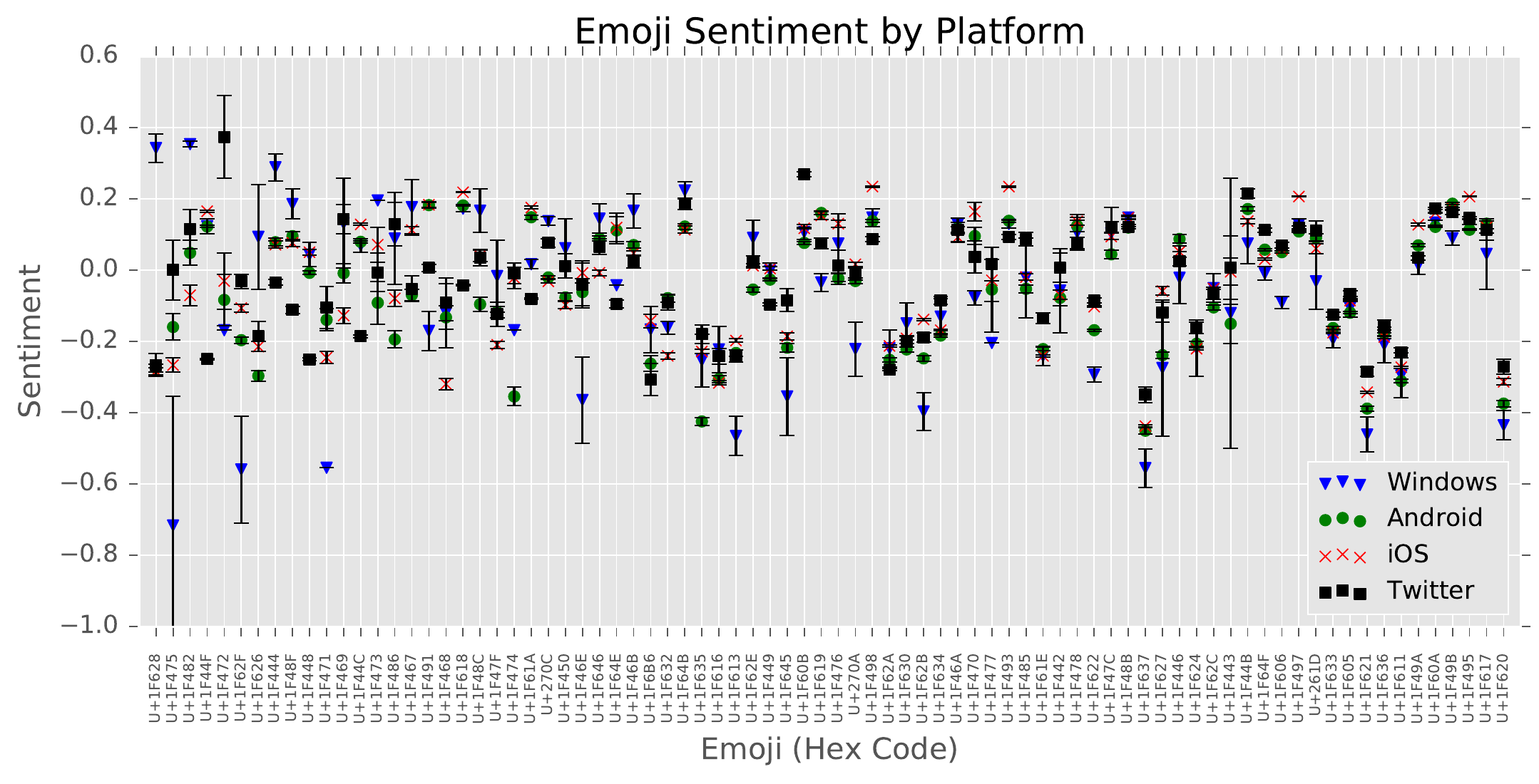}
	\caption{Average sentiment for each platform by emoji. Error bars indicate the variance calculated from the bootstrapped samples. A sentiment score of 0.0 is ``neutral,'' 1.0 is ``perfectly positive,'' and -1.0 is ``perfectly negative.'' The $x$-axis labels indicate the Unicode character code of the emoji. }
	\label{fig:emojidifferences}
\end{figure*}
Word embeddings have the property that words that are more semantically similar will be embedded closer together~\cite{mikolov2013distributed}, where ``closeness'' is defined by cosine similarity. We compare the usage of the emojis on each platform by seeing the words that are embedded closest to each emoji. For each platform, we extract the closest 1,000 words to the emoji. To compare the differences across platforms, we compute the Jaccard coefficient between the top 1,000 on the first platform and the top 1,000 on the second. We compute the average Jaccard coefficient across all emojis.

The results of this experiment are shown in Table~\ref{tab:jaccard}. The results indicate that the emojis are embedded next to very different words across models. The most agreeing platforms are iOS and Android, where an average of 153 words are common across the top 1,000 in the emojis. We also note that Windows has a much lower average agreement than other models. Finally, \emph{all} platforms are extremely different from a random sample. This means that combining tweets from all platforms, as is done in many analytical tasks, will yield a significantly different representation than considering each platform individually. 

While we have discovered that the emojis are used in different contexts across platforms, that does not necessarily mean that their meaning is perceived differently. To better answer this question, we assess the sentiment of the tweets in which the emojis occur.

\subsection{Assessing Emoji Sentiment}
Having collected an emoji dataset, we continue to see if the usage of the emojis is different across platforms. To measure the consistency of the meaning of the emoji, we perform sentiment analysis on the tweets containing the emoji. Using the Pattern library's sentiment analysis tool,\footnote{\url{http://www.clips.ua.ac.be/pages/pattern}} we compute the average sentiment for each emoji on each platform. This is done by removing the emoji from the tweet and using the sentiment analysis tool to compute the sentiment score for the remaining text. Finally, we consider the possibility that each platform may have a different sentiment ``bias,'' that is the sentiment expressed on those platforms is different. For example, Windows Phone may be preferred by business users who are less likely to express negative sentiment in their posts. To account for this, we take the average sentiment across all tweets on the platform, and subtract that from the emoji's score.\footnote{All platforms have roughly the same sentiment bias of +0.20 with the exception of Windows Phone with a sentiment bias of +0.30.} We then take the average sentiment across all tweets in which the emoji occurs and plot the average in Figure~\ref{fig:emojidifferences}. Because we only have one corpus for each platform, we bootstrap~\cite{efron1994introduction} the corpus to obtain confidence intervals. By sampling with replacement, we create 100 bootstrapped samples and reproduce the process above to understand the variance in the data, yielding the confidence intervals in Figure~\ref{fig:emojidifferences}.

The results of this experiment show several phenomena about the usage of emojis online. First, we see that in many of the emojis that there is a significant difference between the sentiment in at least two of the platforms. Among those, some even have \emph{diverging} sentiment polarities. In these cases, if one were to read only the tweets from a particular platform, they would think that emoji has a completely different meaning than it does on another platform. To illustrate this point, we provide an example of the emoji differences. We include their official definition according to the Unicode standard,\footnote{\url{http://unicode.org/emoji/charts/full-emoji-list.html}} and their depiction across the different platforms in Table~\ref{tab:emojidiffs}. Take, for example, the ``fearful face'' emoji. In the case of this emoji, Figure~\ref{fig:emojidifferences} indicates that Android, iOS, and Twitter all have this emoji hovering at the roughly ``neutral'' area of the sentiment spectrum. Windows, however, is an extreme outlier with most users including this emoji in positive tweets. The intuition behind this phenomenon is demonstrated clearly by the ``Fearful Face'' emoji shown in Table~\ref{tab:emojidiffs}, where Android, iOS, and Twitter display the emoji completely differently from Windows. We speculate that the difference in the way the emojis forehead is rendered could cause this difference in interpretation. Another difference that appears in the ``Clapping Hands'' emoji, this time with Twitter being the outlier. Android, iOS, and Windows all clearly show two hands with action lines indicating that they are moving together. Twitter, on the other hand, is less perceptible. Only one hand is clearly visible, and this could give the impression of a ``slap'' motion, yielding the more negative sentiment. In the case of the ``Person Pouting'' emojis, Android and iOS render a significant frown, while Twitter and Windows have a person with a slacked mouth.

\begin{table}
	\caption{Emojis to illustrate the meaning difference. The names and codes are provided by the Unicode standard.}
	\resizebox{1.0\columnwidth}{!}{\begin{tabular}{c | c | c c c c}
	\toprule
	\small{Name} & \small{Code} & \small{Android} & \small{iOS} & \small{Twitter} & \small{Windows} \\
	\midrule
	Fearful Face & U+1F628 & \includegraphics[width=.5cm]{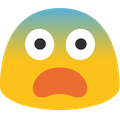} & \includegraphics[width=.5cm]{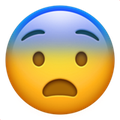} & \includegraphics[width=.5cm]{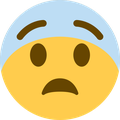} & \includegraphics[width=.5cm]{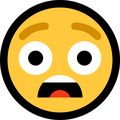} \\
	Clapping Hands & U+1F44F & \includegraphics[width=.5cm]{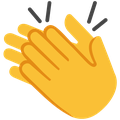} & \includegraphics[width=.5cm]{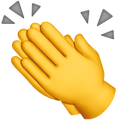} & \includegraphics[width=.5cm]{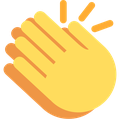} & \includegraphics[width=.5cm]{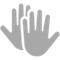} \\
	Person Pouting & U+1F64E & \includegraphics[width=.5cm]{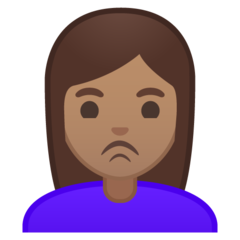} & \includegraphics[width=.5cm]{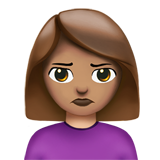} & \includegraphics[width=.5cm]{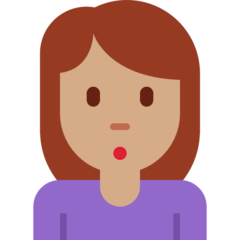} & \includegraphics[width=.5cm]{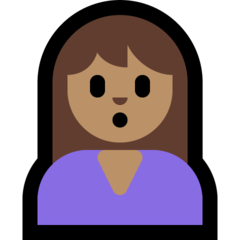} \\
	\bottomrule
	\end{tabular}}
	\label{tab:emojidiffs}
\end{table}

We have now conducted two experiments on the cross-platform use of emojis. These experiments both illustrate that the use of emojis is platform specific, answering \textbf{RQ1}. In the case of our word embedding experiment, we find that the words that neighbor a certain emoji are vastly different between platforms, indicating that they are used in different contexts. This point is furthered by our sentiment analysis experiment, where we found that some emojis have \emph{significantly} different sentiment scores across platforms, confirming the results of~\cite{miller2016blissfully} on a large-scale, real-world dataset.

\subsection{The Scale of Misinterpretation}
\label{sec:scale}
We continue to address \textbf{RQ2}, which involves measuring the scale at which emojis can contribute to miscommunication across platforms. We have used real-world data to show that this problem exists, however, we do not know the extent to which users are affected by this phenomenon. Is this a wide-reaching problem impinging most Twitter users or is it an esoteric issue restricted to the few users who happen to include emojis in their text? 

Based on the results of our experiment, we find that 38.2\% of all emojis yield a statistically significant sentiment difference between different platforms. While this is a minority, these emojis appear in 73.4\% of all of all the tweets in our dataset. Since our dataset was collected using emojis, we need to leverage outside information to estimate the impact for all of Twitter. To estimate the fraction of tweets using these emojis, we use the Sample API,\footnote{https://dev.twitter.com/streaming/reference/get/statuses/sample} which provides a 1\% sample of \emph{all} of the tweets on Twitter, irrespective of whether they contain an emoji. We collected data from the Sample API during the same time period we collected the emoji dataset. Through analyzing this data, we observe that of the 94,233,024 tweets we collected from the Sample API during this time period, 8,129,483 tweets (\textbf{8.627\%}) use emojis that are prone to misinterpretation. In other words, 1 in every 11 tweets sent on Twitter contain an emoji that has a statistically significant interpretation to a user on a different platform.


These findings indicate that there is a need to disambiguate emojis across platforms. To this end, the rest of this paper proposes a strategy for generating a mapping which can disambiguate emoji choice across platforms. Next, we evaluate this mapping to show that it can increase the performance of sentiment analysis tasks.
\begin{figure*}[!ht]
\centering
	\hspace{-1.5em}\subfigure[Platform-Independent Word Embedding]{
		\label{fig:word_embedding}
		\includegraphics[width=0.40\textwidth]{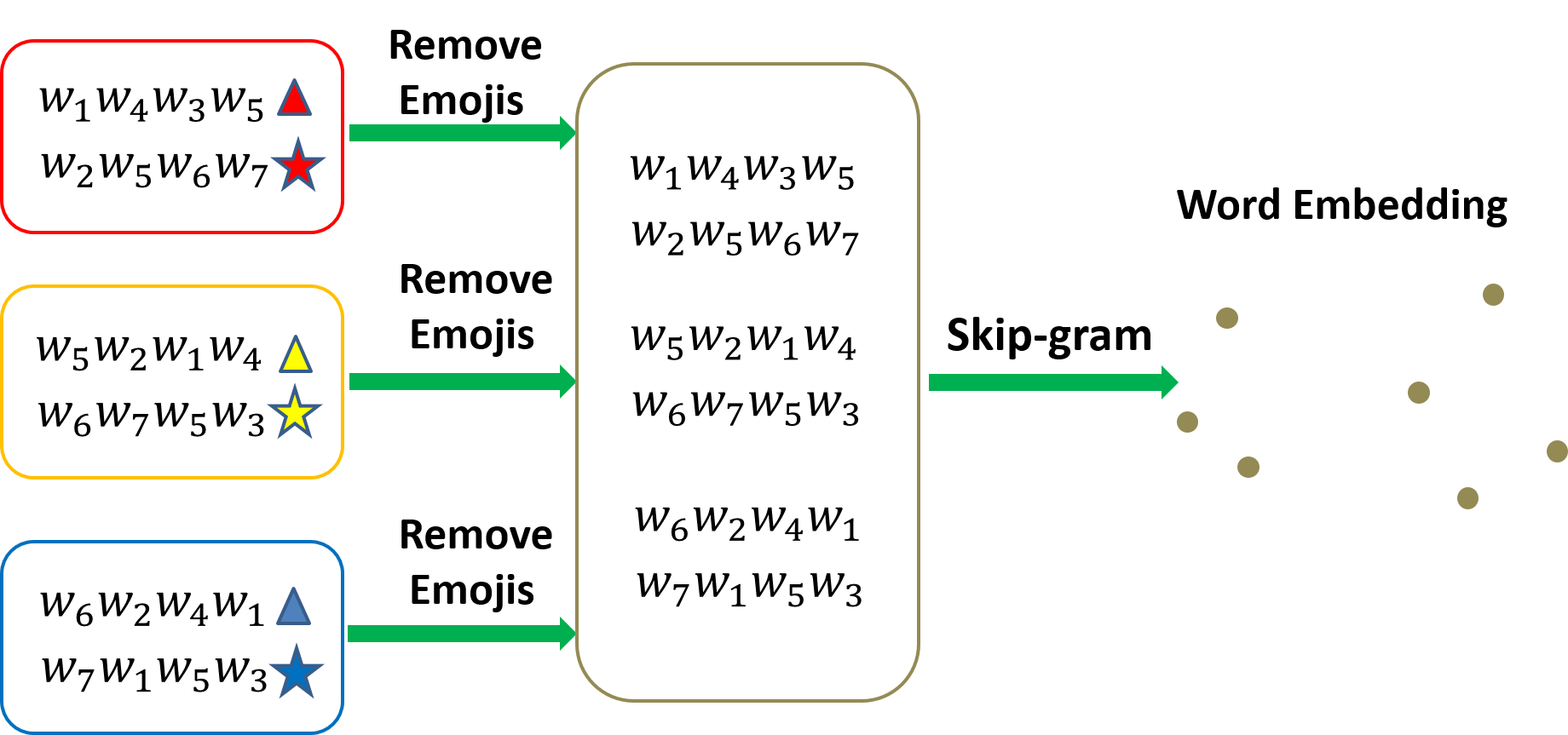}}
	\subfigure[Platform-Dependent Emoji Embedding]{
		\label{fig:emoji_embedding}
		\includegraphics[width=0.40\textwidth]{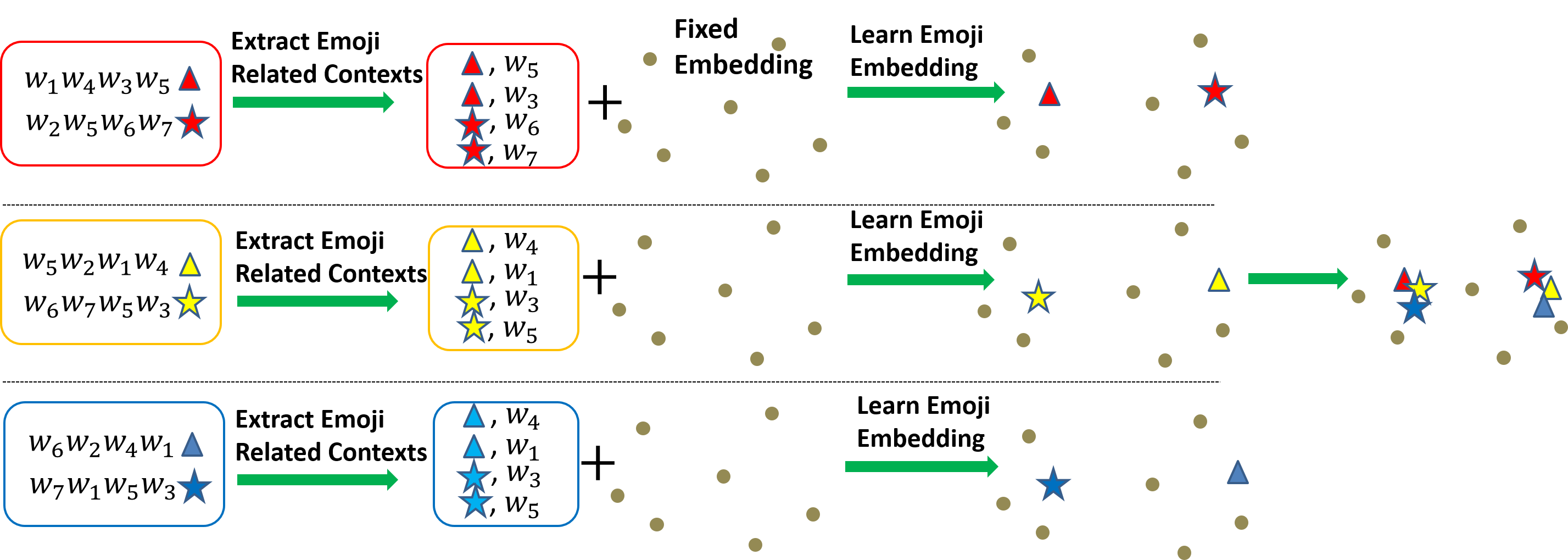}}
	\caption{An illustration of platform-dependent emoji mapping construction. The ``Word Embedding'' that is output at the end of Figure~\ref{fig:word_embedding} is then used as the ``Fixed Embedding'' in Figure~\ref{fig:emoji_embedding}. The three rectangles blue, green, red means corpus of three platforms. Each line is a sentence. Triangle and Star denote emojis. Red triangle means an emoji in red platform while yellow triangle means the same emoji in yellow platform.} \label{fig:dict_construction_flow_chart}
\end{figure*}

\section{An Emoji Mapping Solution}
The fact that the representations of emojis are different both from the perspective of the embedding as well as the sentiment suggests that a mapping may help us disambiguate emoji meaning across platforms. In this section, we describe our approach to constructing a platform-dependent emoji mapping, addressing \textbf{RQ3}. The goal of this mapping is to provide a translation from an emoji on one platform to its corresponding emoji on another. The purpose of the platform-dependent emoji mapping is to disambiguate platform-specific emoji interpretation. To construct the cross-platform emoji mapping, we need to understand the semantic meaning and the sentiment polarity of the emojis on each platform. We then can construct the platform-dependent mapping by identifying emojis that have the closest semantic and sentiment polarities across platforms. The key step is to learn the representation of emojis from the texts on each platform such that the representations capture the platform-dependent semantic meanings of emojis and also allow for similarity matching to construct the mapping.

Recent advances suggest that word embeddings such as skip gram~\cite{mikolov2013distributed} and GloVe~\cite{pennington2014glove} are able to capture the semantic meanings of words. The low-dimensional vector representations also allow a similarity calculation using cosine or Eulidean distance, which eases the mapping construction. In addition, recent findings on emoji analysis~\cite{barbieri2016does,eisner2016emoji2vec} demonstrate that by treating each emoji as a word and performing skip gram on texts, the vector representation learned by skip gram can capture semantic meanings and sentiment polarities of emojis, which improve the performance of sentiment classification.

Word embeddings cannot be directly applied to solve this problem. There are two challenges that we must overcome in order to create the mapping. The first challenge is that we must find a way to map emojis from a source platform to their true equivalent semantic emoji on the target platform. The second challenge is that when we build our platform-specific embedding, the position of the \emph{words} will change, as well as the emojis. Thus, we need to figure out a way to represent the words within a common space first, so that we can extend it to measure the emojis relative to the words. Our solution addresses both of these challenges.


\subsection{Building the Embedding}
\label{sec:mappingsolution}
For simplicity, let $\mathcal{T}_p$, $p=1,\dots,P$ be the set of tweets from each platform. The process of building word and emoji embeddings are illustrated in Figure \ref{fig:dict_construction_flow_chart}. As shown in Figure \ref{fig:word_embedding}, since the interpretation of words are platform-dependent, for each corpus $\mathcal{T}_p$, we first remove the emojis, which are denoted as $\tilde{\mathcal{T}}_p$. We then combine $\tilde{\mathcal{T}}_p$, $p=1,\dots,P$, as one large corpus $\tilde{\mathcal{T}}$. Then skip gram is applied on $\tilde{\mathcal{T}}$ to learn word embedding $\mathbf{W} \in \mathbb{R}^{K \times N}$, where $K$ is dimension of the vector representation and $N$ is the size of the vocabulary without emojis. The advantages of combining $\tilde{\mathcal{T}}_p$, $p=1,...,P$, as one large corpus $\tilde{\mathcal{T}}$ are two-fold: (i) we obtain a large corpus which allows us to train a better word embedding; and (ii) word embedding in each platform is the same, which satisfies the assumption that word interprations are the same for each platform. After the platform-independent word embedding is learned, then we can learn the platform-dependent embeddings for emojis. The process is depicted in Figure \ref{fig:emoji_embedding}. For each corpus $\mathcal{T}_p$, as the process of skip gram, we first use a context window of size 5 to extract the neighboring words of emojis in the corpus. Let $(e_i,w_j)$ denote a pair of neighboring words, where $e_i$ means emoji $i$ and $w_j$ means word $j$. The extracted pairs are then put into the set $\mathcal{P}_p$. We then learn the representation of emoji $e_i$ in corpus $\mathcal{T}_p$ by optimizing the following problem:
\begin{equation}
\label{eq:emoji_obj}
	\min_{\mathbf{e}_i^{p}} \sum_{w_j: (e_i,w_j)\in \mathcal{P}_p} \Big( \log \sigma (\mathbf{w}_j^T \mathbf{e}_i^{p}) + \sum_{k=1}^{N} \log \sigma(-\mathbf{w}_k^T \mathbf{e}_i^p) \Big)
\end{equation}
where $\mathbf{e}_i^p$ is the vector representation of $e_i$ in corpus $\mathcal{T}_p$. Equation \ref{eq:emoji_obj} is essentially the negative sampling form of the objective function of the skip gram approach, where we try to learn the emoji representaion which is able to predict the neighboring words. Note that the difference with skip gram is that word embeddings $\mathbf{W}$ is fixed across different corpus, we only learn $\mathbf{e}_i^p$. We do the same thing for each corpus, which gives us $\mathbf{e}_i^p$, $p=1,\dots,P$, i.e., the vector representations of the same emoji in different platforms.

\subsection{Constructing the Mapping}
In this section, we detail the emoji mapping construction process. To construct the mapping between emojis across different platforms, we consider it in a pair-wise scenario. We treat one platform as the source platform and the other as the target platform. Without loss of generality, let $\mathcal{E}=\{e_i, i \in \{1,...,m\}\}$ be the set of emojis that occur in both platforms. By learning the emoji embedding representations in each platform, we are able to capture the platform-dependent semantic features for the emojis. Thus, given an emoji in the source platform, we can leverage the emoji embedding representation to connect the semantic space between the source and target platforms, and then find the most similar emoji in the target platform.

Specifically, based on all the emoji embeddings from the source platform $\{\mathbf{e}_i^{s}, i\in\{1,...,m\}\}$ and target platform $\{\mathbf{e}_i^{t}, i\in\{1,...,m\}\}$, we want to map the most similar emoji in target platform for each emoji in the source platform.
To compute the similarity of two emoji embeddings, we adopt the cosine similarity measure as follows,
\begin{equation}
\label{eq:emoji_sim}
	sim(\mathbf{e}_i^{s},\mathbf{e}_j^{t}) = \frac{\mathbf{e}_i^{s} \cdot \mathbf{e}_j^{t}}{||\mathbf{e}_i^{s}|| \cdot ||\mathbf{e}_j^{t}||}
\end{equation}

Given an emoji $e_i$ in the source platform, we first compute the similarity between $e_i$ with all emojis in target platform. Then we select the emoji which gives the maximum similarity score. We solve the following objective function to obtain the mapping emoji $\hat{e_j}$ in target platform for $e_i$,
\begin{equation}
\label{eq:emoji_map}
	\hat{e_j} = \arg\max_{e_j\in\mathcal{E}} sim(\mathbf{e}_i^{s},\mathbf{e}_j^{t})
\end{equation}
we then get a mapping pair $(e_i,\hat{e_j})$, where the first emoji is from the source platform and the second one is from the target platform. Note that the emoji mappings are directional, which means if $(e_i, \hat{e_j})$ is a mapping pair, $(\hat{e_j},e_i)$ is not necessarily a mapping pair.

\section{Evaluating the Mapping}
Now that we have presented the methodology for constructing the emoji mapping, we will validate its utility, answering \textbf{RQ4}. We measure the utility of our mapping by seeing how well it can help in a common text analysis task: sentiment analysis. We design experiments to show that the sentiment analysis task is improved by applying our emoji mapping to the data to bolster the consistency of analogy meaning in the dataset. This simultaneously shows the efficacy of our approach as well as the extent to which emoji ambiguity plays in standard text analysis tasks.



\subsection{Predictive Evaluation}
We have shown that emoji usage is different across platforms. This means that the same emoji can have different sentiment meanings across platforms. From a machine learning perspective, this means that platform-specific emoji renderings introduce \emph{noise} into the dataset, ultimately lowering the classification performance. We hypothesize that by applying our mapping we are in effect converting the dataset into a single language emoji dataset, and the consistent emoji language will help in assigning a sentiment label.
\begin{figure*}
	\hspace{-3em}\subfigure[Accuracy]{
		\includegraphics[height=.3\textheight]{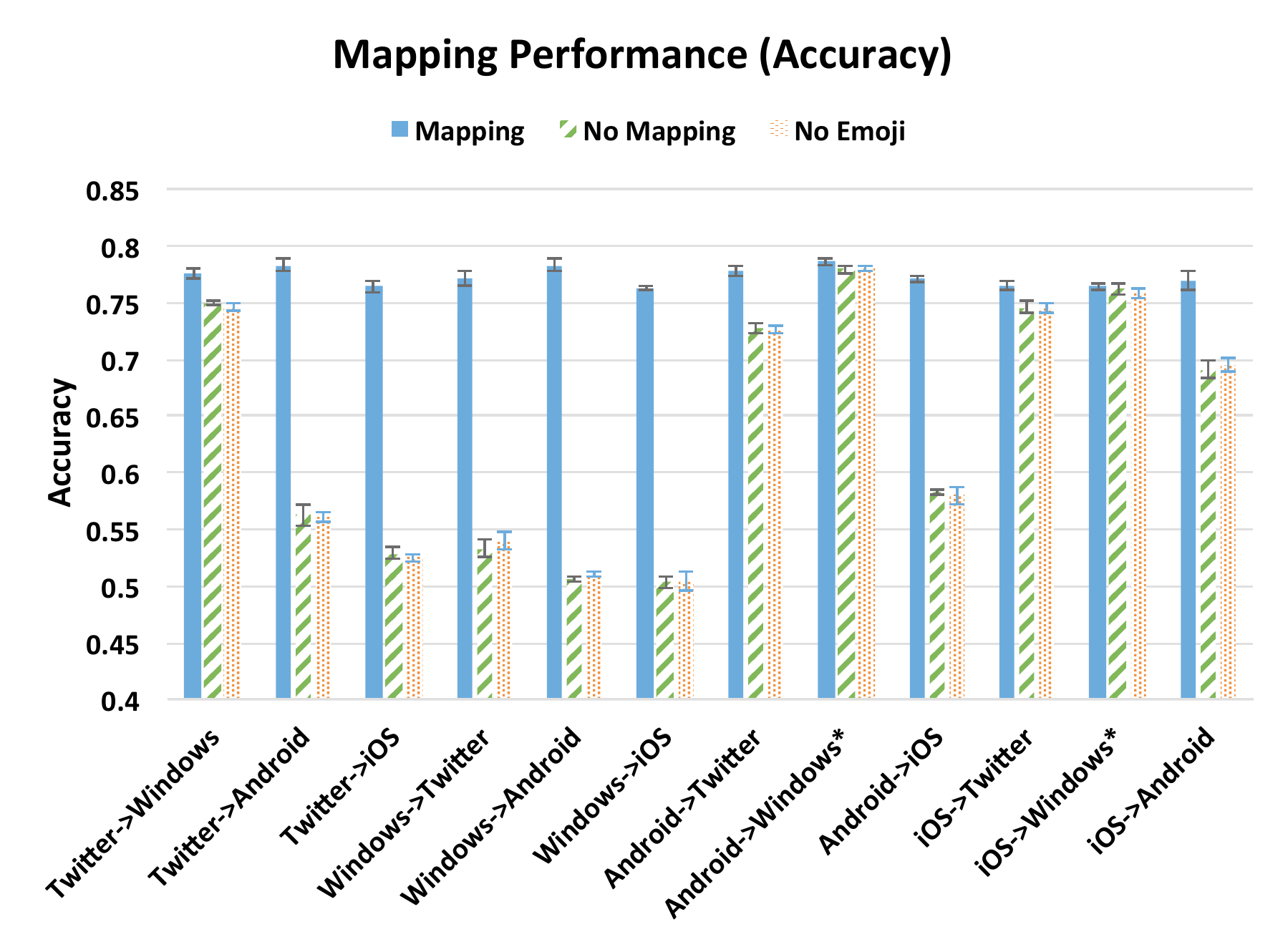}	
		\label{fig:accuracyplot}
	}
	\subfigure[F1]{
		\includegraphics[height=.3\textheight]{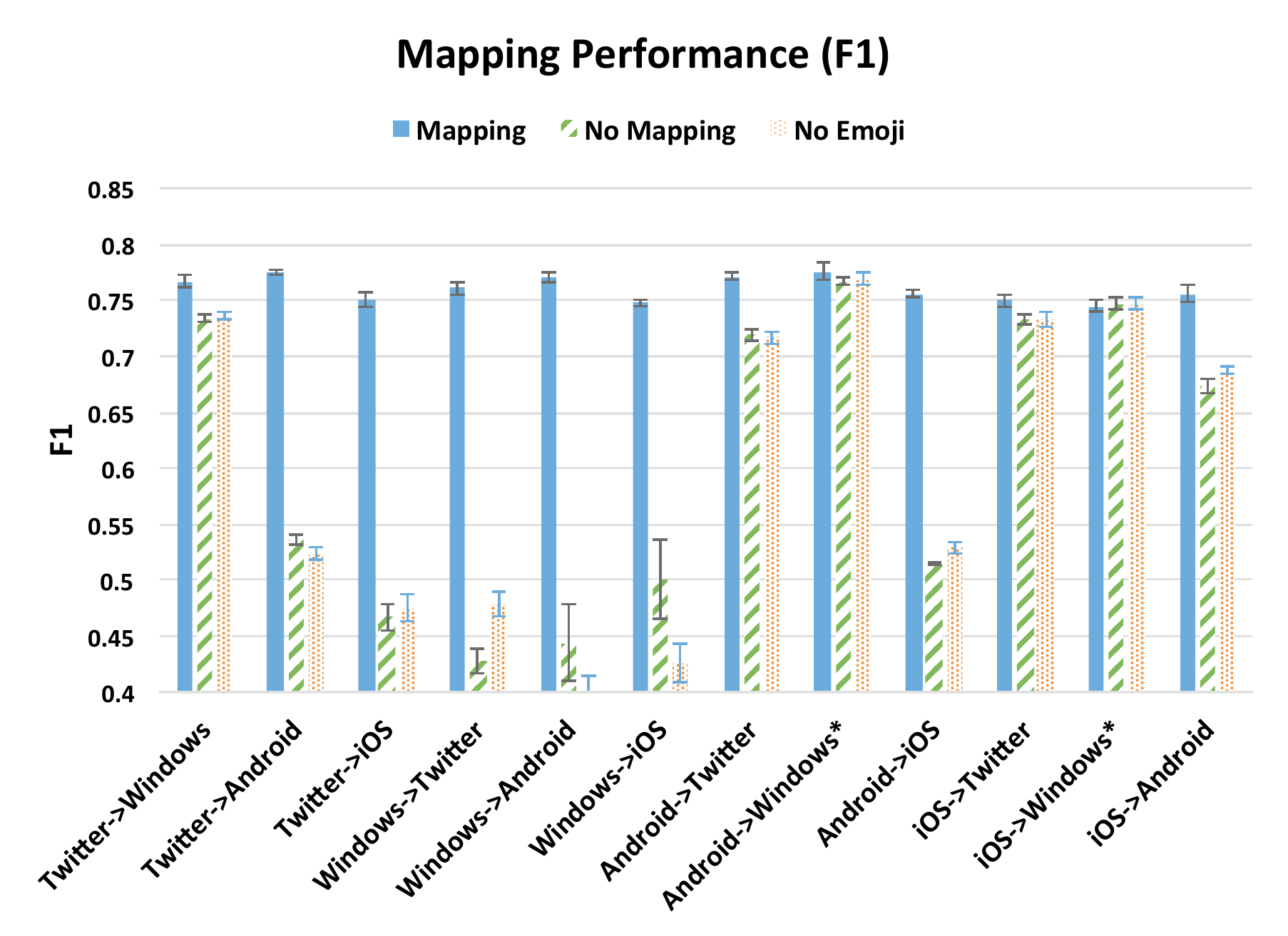}	
		\label{fig:f1plot}
	}
	\caption{Performance results across all ``source'' X ``target'' pairs. Asterisks indicate (source, target) pairs where the mapping is \emph{not} significantly better than the two baselines. F$_1$ is computed with respect to the ``positive'' class. These results are taken when deleting tweets with sentiment scores in the range of (-0.2, 0.2).}
	\label{fig:perf}
\end{figure*}

\subsection{Experimental Setup}
In each experiment, we evaluate our mapping using a pair of platforms. One of the platforms is the ``source'', and another is the ``target.'' The experiment consists of two phases. In the first phase, we mix the data from both platforms together, perform sentiment analysis, and measure the accuracy using 5-fold cross validation. We call the average accuracy across all 5 folds ``A1.'' In phase two, we apply the mapping to the emojis on the ``target'' platform, and then repeat the process in phase one. We call the average accuracy from this experiment ``A2.'' We measure the effectiveness of the mapping as A2 - A1.

In preprocessing, we remove stopwords, and strip the case from all words. We tokenize the dataset using the ``TweetTokenizer'' module in Python's NLTK~\cite{bird2009natural}. Following the labeling approach outlined previously, we use the Pattern library to assign a sentiment score to each tweet as ground truth, again hiding the emojis. The sentiment score provided by this library is a continuous value from [-1.0, 1.0]. We convert this problem to a binary classification task. We delete all tweets in the range (-0.2, 0.2) to ignore ambiguous cases. We then assign the sign of the label from the Pattern library as the label of the tweet. When training the word embeddings as well as the emoji embeddings, we use $K=20$ as the number of dimensions.

To prevent information leakage between the training and test sets, the emoji mapping used in these experiments is built using data collected from September 23rd, 2016 - October 4th, 2016. The training and test instances used in our cross validation experiments are taken from October 5th, 2016 - October 20th, 2016.

Each tweet is represented by the average of the vectors for its words $\mathcal{W'}$ and emojis $\mathcal{E'}$ in the tweet. We use the target embedding for emojis. Thus, a tweet is represented as follows:
\begin{equation}
	repr= \frac{1}{|\mathcal{W'}|} \sum_{w_k \in \mathcal{W'}} \mathbf{w}_k + \frac{1}{|\mathcal{E'}|} \sum_{e_i \in \mathcal{E'}} \mathbf{e}_i^t,
	\label{eqn:nomap}
\end{equation}

where $\mathbf{w}$ and $\mathbf{e}^t$ are the source embedding and target embedding. When the mapping is applied, the representation is as follows:
\begin{equation}
	repr\_map= \frac{1}{|\mathcal{W'}|} \sum_{w_k \in \mathcal{W'}} \mathbf{w}_k + \frac{1}{|\mathcal{E'}|} \sum_{e_j \in \mathcal{E'}} \mathbf{\hat{e}}_j^t,
	\label{eqn:map}
\end{equation}
where $\hat{e}_j$ is the mapping of the target emoji on the source platform.

Having extracted the data, the labels, and formalized the representation, we use SVM to build a classifier using 5-fold cross validation. The only difference between each set up is how each tweet is represented. We compare three tweet representations:
\begin{enumerate}
	\item \textbf{Mapping.} This is the representation where the emojis in the target platform are mapped to the emojis in the source platform. This corresponds to the representation obtained by Equation~\ref{eqn:map}.
	\item \textbf{No Mapping.} This is the representation when no mapping is applied obtained by Equation~\ref{eqn:nomap}.
	\item \textbf{No Emojis.} Our hypothesis is that the incorrect emojis are adding noise to our dataset, which hinders classification. Aside from our proposed solution, another way to de-noise the dataset is to simply remove the emojis. We do this by using the following representation formula:
	\begin{equation}
		repr\_noemoji = \frac{1}{|\mathcal{W'}|} \sum_{w_k \in \mathcal{W'}} \mathbf{w}_k.
	\end{equation}
	In this way, emojis are simply not considered in the resulting feature representation.
\end{enumerate}

\vspace{1em}
\subsection{Experimental Results}
\label{sec:rawresults}
We report the average across the 5 folds in Figure~\ref{fig:perf}. The results overall are encouraging. In most of the source/target pairs we obtain significantly better results than both doing nothing, and removing the emojis. Further, we note that removing the emojis beats doing nothing in almost all of the cases, further validating our noise assumption. The only two pairs where the mapping does not obtain significantly better results are iOS $\rightarrow$ Windows (``source'' $\rightarrow$ ``target''), and Android $\rightarrow$ Windows. The lower results across these two cases could be a side effect of the nature of the Windows emojis. The results of our previous analysis indicate that the Windows emoji set is significantly different from the rest in many cases, and that could prevent us from learning a quality mapping.


Using the random 1\% sample of Twitter data introduced in Section~\ref{sec:scale}, we discover that 15.0\% of all tweets contain at least one of any emoji. Our analysis from the same section indicates that 8.627\% of all tweets, and thus 57.5\% of tweets containing an emoji are affected by this phenomenon. This justifies the huge improvements seen in Figure~\ref{fig:perf}. We speculate that another factor that contributes to the superior performance is the large amount of data upon which the embeddings were trained. These embeddings are available online.\footnote{\url{http://www.public.asu.edu/~fmorstat/emojimapping/}}

\subsection{Results by Sentiment Threshold}
\label{sec:sentisignif}
In the previous experiment we removed tweets that had a sentiment score between (-0.2, 0.2). The motivation behind this step is that tweets within this threshold were so ambiguous that they could not be meaningfully assessed for sentiment. In this section, we vary this parameter to see how robust our method is to ambiguous tweets. We vary the sentiment threshold from 0.1 (leaving many ambiguous tweets) to 0.9 (leaving only the most sentiment-expressive tweets).

Instead of showing the results of every possible combination, which is impossible due to space limitations, we instead test whether the two sets of results differ significantly. The T-test tests the null hypothesis that the means of the two distributions are equal. The results can be seen in Table~\ref{tab:ttest_nomapping} for the comparison between Mapping and No Mapping, and in Table~\ref{tab:ttest_noemoji} for the comparison between Mapping and No Emoji. In the first case we can easily reject this null hypothesis at the $\alpha = 0.05$ significance level in all cases except for iOS$\rightarrow$Windows. Despite passing the significance bar, this is pair still yields the least significant result in Table~\ref{tab:ttest_noemoji}. This is consistent with our previous result, where this particular mapping did not fare significantly better, and further supports our suspicion that Windows is an outlier due to the way it renders emojis.

\begin{table}[t]
	\centering
	\caption{Sentiment threshold significance T-test between ``Mapping'' and ``No Mapping'' experimental designs.}
	\begin{tabular}{lcc}
		\toprule
		& \textbf{Accuracy} & \textbf{F1} \\
		\midrule
		Twitter$\rightarrow$Windows & 3.538e-09 & 1.200e-05\\
		Twitter$\rightarrow$Android & 2.493e-08 & 8.754e-07\\
		Twitter$\rightarrow$iOS & 8.071e-08 & 4.887e-07\\
		Windows$\rightarrow$Twitter & 4.418e-07 & 5.687e-05\\
		Windows$\rightarrow$Android & 3.538e-09 & 8.440e-07\\
		Windows$\rightarrow$iOS & 2.115e-08 & 1.960e-05\\
		Android$\rightarrow$Twitter & 1.234e-07 & 1.193e-05\\
		Android$\rightarrow$Windows & 2.586e-07 & 3.037e-03\\
		Android$\rightarrow$iOS & 2.081e-04 & 6.212e-06\\
		iOS$\rightarrow$Twitter & 2.726e-06 & 1.313e-05\\
		iOS$\rightarrow$Windows & 0.294 & 0.094\\
		iOS$\rightarrow$Android & 8.877e-08 & 3.407e-06\\
		\bottomrule
	\end{tabular}
	\label{tab:ttest_nomapping}
\end{table}

\begin{table}[t]
	\centering
	\caption{Sentiment threshold significance T-test between ``Mapping'' and ``No Emoji'' experimental designs.}
	\begin{tabular}{lcc}
		\toprule
		& \textbf{Accuracy} & \textbf{F1} \\
		\midrule
		Twitter$\rightarrow$Windows & 2.466e-06 & 5.142e-08\\
		Twitter$\rightarrow$Android & 8.818e-09 & 1.923e-09\\
		Twitter$\rightarrow$iOS & 6.121e-08 & 2.400e-08\\
		Windows$\rightarrow$Twitter & 6.047e-08 & 1.665e-07\\
		Windows$\rightarrow$Android & 3.293e-08 & 3.141e-07\\
		Windows$\rightarrow$iOS & 9.665e-08 & 1.417e-06\\
		Android$\rightarrow$Twitter & 3.525e-06 & 2.021e-06\\
		Android$\rightarrow$Windows & 1.744e-03 & 1.124e-04\\
		Android$\rightarrow$iOS & 7.423e-08 & 8.920e-09\\
		iOS$\rightarrow$Twitter & 2.078e-05 & 2.954e-06\\
		iOS$\rightarrow$Windows & 3.701e-04 & 3.399e-02\\
		iOS$\rightarrow$Android & 3.303e-08 & 4.338e-08\\
		\bottomrule
	\end{tabular}
	\label{tab:ttest_noemoji}
\end{table}

\vspace{1em}
\section{Conclusions and Future Work}
\label{sec:conclusions}
In this work, we set out to answer a series of questions regarding the nature and extent of cross-platform emoji misinterpretation, and to provide a solution that can help researchers and practitioners to overcome this platform-specific inconsistency in their analysis. In the introduction, we outlined four research questions based around these issues. Here, we summarize our findings as they pertain to each question and end with a discussion of areas of future work.

\emph{\textbf{RQ1} Does emoji misinterpretation occur in the real world?} To perform this study we use Twitter, a large, open platform where users post hundreds of millions of tweets per day, and where the data made available by Twitter contains the source from which the tweet was posted, which can be used to identify the underlying platform. We find that the sentiment of the tweets in which emojis occur differs widely and significantly across platforms. In many cases, the same emoji exhibited opposing sentiment polarities on different platforms. 

\emph{\textbf{RQ2} What is the scale of this misinterpretation?} By analyzing a random sample of tweets, we obtain that 8.627\%, or roughly 1 in every 11 of all of tweets contain an emoji that is used in a statistically significantly different fashion on a different platform.

\emph{\textbf{RQ3} How can the problem of cross-platform emoji interpretation be addressed?}
With these findings, we endeavor to construct an emoji mapping to help researchers, practitioners, and even readers of social media data to better understand the intended message of the sender given their platform-specific emoji mapping. We construct a solution that exploits the property that ``similar words are embedded closer together'' in order to find the corresponding emoji across platforms and build a mapping.

\emph{\textbf{RQ4} Does correcting for emoji misinterpretation have a meaningful effect on analysis?}
We evaluate the effectiveness of our embedding by applying it to sentiment analysis. We chose sentiment analysis as it is a prediction problem very common to social media data. We show that by mapping all tweets to a consistent emoji vocabulary, we can significantly increase the performance of sentiment analysis by diminishing the amount of emoji noise in the dataset.

This work opens up the doors for many areas of research. While we have largely explored this platform disagreement in emoji meanings from the perspective of computational and predictive tools, it certainly is not limited to this. In fact, it may be useful to include our mapping in user interfaces in order to increase understanding between individuals communicating on different platforms. Alternatively, social media platforms can take the approach of a ``closed'' platform, where all platforms conform to a single emoji set. WhatsApp is an exemplar of this closed approach.

The mapping, and the data used in this work are shared in accordance with Twitter's data sharing policies online.\footnote{\url{http://www.public.asu.edu/~fmorstat/emojimapping/}} We also provide all of the performance scores used to obtain the significance results from the sentiment threshold experiment, as well as an expanded version of Figure~\ref{fig:emojidifferences} containing more emojis.

\bibliographystyle{ACM-Reference-Format}
\bibliography{sigproc}

\end{document}